\documentclass[a4paper]{article}

\usepackage[english]{babel}
\usepackage[utf8]{inputenc}

\usepackage{INTERSPEECH2021}
\usepackage{cite}
\usepackage{amsfonts}
\usepackage{amsmath}
\usepackage{xcolor, graphicx}
\usepackage{booktabs}
\usepackage{multirow}
\usepackage{tabularx}

\usepackage{units}  
\usepackage{siunitx,etoolbox}
\usepackage[capitalize]{cleveref}
\Crefname{equation}{Eq.}{Eqs.}
\Crefname{figure}{Fig.}{Figs.}
\Crefname{tabular}{Tab.}{Tabs.}

\DeclareMathOperator*{\argmax}{arg\,max \hspace{2mm}}

\newcommand\numberthis{\addtocounter{equation}{1}\tag{\theequation}}

\newcommand{\blank}{{\epsilon}}

\title{Equivalence of Segmental and Neural Transducer Modeling:\\ A Proof of Concept}
\name{Wei Zhou$^{1,2}$, Albert Zeyer$^{1,2}$, André Merboldt$^{1}$, Ralf Schl\"uter$^{1,2}$, Hermann Ney$^{1,2}$}
\address{
$^1$Human Language Technology and Pattern Recognition, Computer Science Department,\\
  RWTH Aachen University, 52074 Aachen, Germany\\
$^2$AppTek GmbH, 52062 Aachen, Germany}
\email{\{zhou, zeyer, schlueter, ney\}@cs.rwth-aachen.de, andre.merboldt@rwth-aachen.de}

\begin{document}

\maketitle
\begin{abstract}
  With the advent of direct models in automatic speech recognition
  (ASR), the formerly prevalent frame-wise acoustic modeling based on
  hidden Markov models (HMM) diversified into a number of modeling
  architectures like encoder-decoder attention models, transducer
  models and segmental models (direct HMM). While transducer models
  stay with a frame-level model definition, segmental models are
  defined on the level of label segments directly. While
  (soft-)attention-based models avoid explicit alignment, transducer
  and segmental approach internally do model alignment, either by
  segment hypotheses or, more implicitly, by emitting so-called blank
  symbols. In this work, we prove that the widely used class of
  RNN-Transducer models and segmental models (direct HMM) are
  equivalent and therefore show equal modeling power. It is shown that
  blank probabilities translate into segment length probabilities and
  vice versa. In addition, we provide initial experiments
  investigating decoding and beam-pruning, comparing time-synchronous
  and label-/segment-synchronous search strategies and their
  properties using the same underlying model.
\end{abstract}

\section{Introduction}

Most acoustic models in speech recognition can be categorized into
being defined on the level of each time frame or directly on the level
of labels and/or segments.  The hybrid neural network (NN) - hidden
Markov model (HMM) \cite{bourlard1994hybrid,zeyer17:lstm} and
extensions \cite{raissi2020interspeech}, CTC \cite{graves2016ctc} and
their generalized transducer variants
\cite{graves2012sequence,sak2017rna,%
  tripathi2019monoRNNT,variani2020hat,%
  zeyer2020transducer,zhou2021phonemeTransducer} are all
defined on the time-frame level.  The attention-based encoder-decoder model
\cite{sutskever2014seq2seq,bahdanau2015nmt,%
  zeyer2018attention,park2019specaugment,tuske2020swbatt} and in
general segmental models
\cite{ostendorf1996segmental,zweig2009segmental,%
  beck2018:segmental,beck2018:alignment, zeyer2021latentatt} are  
defined on a label and/or segment level.  Segmental models generalize
from the encoder-decoder model by introducing an explicit latent
variable which usually represents the temporal position of a label
\cite{zeyer2021latentatt}.  Such explicit representation of the time
is needed to enable monotonicity and to potentially allow for online
streaming. 
While \cite{variani2020hat} briefly mentioned the duration interpretation of the blank probability of a transducer model, the general interrelation between time-synchronous models and segmental models are not well studied in the literature.

In this work, we prove the equivalence of transducer models and
segmental models, with the latter introducing an explicit
representation for the temporal position or boundaries per label
segment as a latent variable. This equivalence implies that we can use
both label-synchronous or time-synchronous beam search decoding for
either case.  We provide initial experiments comparing
label-synchronous and time-synchronous beam search decoding for the
transducer models introduced in
\cite{zhou2021phonemeTransducer,zeyer2020transducer}.

\section{Model Equivalence}
\label{sec:sec2}
Let $x_1^{T'}$ denote the input feature sequence and
$h_1^T = f^{\text{enc}}(x_1^{T'})$ denote the encoder output, which
transforms the input into a sequence of high-level representations.
In general, $T \le T'$ due to optional sub-sampling in the encoder. Let $a_1^S \in V^S$ denote a label sequence of length $S$ from a vocabulary $V$.
The general sequence-to-sequence models target at the following sequence posterior probability:\\
\scalebox{0.9}{\parbox{1.11\linewidth}{%
\begin{align*}
p(a_1^S \mid x_1^{T'}) = p(a_1^S \mid h_1^T) \numberthis \label{eq:seqPosterior}
\end{align*}}}
In the following, we show the equivalence between segmental modeling and transducer modeling of \Cref{eq:seqPosterior} under the constraint that all $T$ encoder output frames have to be consumed (denoted as constraint-$T$). 
Without loss of generality, we show that a transducer model based on the RNN-T topology \cite{graves2012sequence} can be rewritten into a segmental model which allows zero-frame segments, and also that a segmental model can be rewritten into an RNN-T model. For both rewriting directions, we provide the corresponding transformation equations. Also the case of assuming transducer models with strict monotonicity as in \cite{sak2017rna, tripathi2019monoRNNT}, is covered as a special case of the equivalence, which leads to a segmental model with minimum segment length of one frame. 


\subsection{Segmental Model}\label{sbs:segmentalmodel}
In a segmental model, the label sequence posterior from \Cref{eq:seqPosterior} can be formulated as:\\
\scalebox{0.9}{\parbox{1.11\linewidth}{%
\begin{align*}
p(a_1^S \mid h_1^T) &= \sum_{t_1^S} p(a_1^{S+1}, t_1^{S+1} \mid h_1^T) \numberthis \label{eq:segSeq}\\
&= \sum_{t_1^S} \prod_{s=1}^{S+1} p(a_s, t_s \mid a_1^{s-1}, t_1^{s-1}, h_1^T) \\
&= \sum_{t_1^S} \prod_{s=1}^{S+1} p(t_s \mid a_1^{s-1}, t_1^{s-1}, h_1^T) \cdot p(a_s \mid a_1^{s-1}, t_1^s, h_1^T)
\end{align*}}}
where $t_s$ are the segment boundaries of label $a_s$ under the monotonicity constraints $t_{s-1} \le t_s \le T$ for $1 \le s \le S+1$.
Here we explicitly introduce a final sentence end label $a_{S+1}=\#$, and define $t_0 \equiv 1$ and $t_{S+1} \equiv T$ for the constraint-$T$.

\subsection{Transducer Model}
Within a transducer modeling approach using the RNN-T topology, \Cref{eq:seqPosterior} can be formulated as:\\
\scalebox{0.9}{\parbox{1.11\linewidth}{%
\begin{align*}
  p(a_1^S \mid h_1^T)
  &= \sum_{y_1^U:a_1^S} p(y_1^U \mid h_1^T) \\
  &= \sum_{y_1^U:a_1^S} \prod_{u=1}^{U=T+S} q(y_u | y_1^{u-1}, h_1^T) \numberthis \label{eq:transducerSeq}
\end{align*}}}
where $y_1^U$ is the blank $\epsilon$-augmented alignment sequence of $a_1^S$ and $q$ is a probability distribution defined over $\bar{V}=V\cup \{\blank\}$. The RNN-T label topology is composed of horizontal transitions for $y_u=\epsilon$ and vertical transitions for $y_u \neq \epsilon$, where a final blank transition at step $U$ is always presented for termination.

\subsection{From Transducer to Segmental Model}
\label{sec:t2s}
We postulate that any transducer model using the RNN-T topology as
defined in Eq.~(\ref{eq:transducerSeq}) can be rewritten into an
equivalent segmental model that defines exactly the same overall label
posterior distribution. The following transformation equations show
the resulting segmental model written in terms of the transducer
model:\\
\scalebox{0.9}{\parbox{1.11\linewidth}{%
\begin{align*}
  p(t_s \mid a_1^{s-1}, t_1^{s-1}, h_1^T) = &\prod_{t=t_{s-1}}^{t_s-1} q(y_{t+s-1}=\blank \mid y_1^{t+s-2}, h_1^T) \\
  & \cdot (1-q(y_{t_s+s-1}=\blank \mid y_1^{t_s+s-2}, h_1^T)) \numberthis \label{eq:durationProb}
\end{align*}}}
and\\
\scalebox{0.9}{\parbox{1.11\linewidth}{%
\begin{align*}
  p(a_s \mid a_1^{s-1}, t_1^s, h_1^T) = \frac{q(y_{t_s+s-1}=a_s \mid y_1^{t_s+s-2}, h_1^T)}{1 -q(y_{t_s+s-1}=\blank \mid y_1^{t_s+s-2}, h_1^T)}  \numberthis \label{eq:labelProb}
\end{align*}}}
with the alignment sequence $y_1^{U}$ defined as follows:\\
\scalebox{0.9}{\parbox{1.11\linewidth}{%
\begin{equation*}
  y_u=\left\{
    \begin{array}{ll}
      a_s & \text{iff} \; \exists\; s: u=t_s+s-1 \\
      \blank & \text{otherwise}
    \end{array}
    \right. \; \forall\; u=1,\ldots,T+S.
\end{equation*}}}
Separately, we compute the probability for sentence end in the segmental model as follows:\\
\scalebox{0.9}{\parbox{1.11\linewidth}{%
\begin{align*}
  p(a_{S+1}=\# \mid a_1^{S}, t_1^{S+1}, h_1^T) = \frac{q(y_{T+S}=\blank \mid y_1^{T+S-1}, h_1^T)}{1 -q(y_{T+S}=\blank \mid y_1^{T+S-1}, h_1^T)}  \numberthis \label{eq:sentendProb}.
\end{align*}}}
Similar as in \cite{beck2018segmentalLVCSR}, the segment duration probability in \Cref{eq:durationProb} is represented as a Bernoulli-like length distribution by regarding $\epsilon$ as a pooled state for segment continuation.
Substituting Eqs.~(\ref{eq:durationProb}) and (\ref{eq:labelProb}) into the segmental model defined in Eq.~(\ref{eq:segSeq}) leads to the transducer model defined in Eq.~(\ref{eq:transducerSeq}).
This shows that using Eqs.~(\ref{eq:durationProb}) and (\ref{eq:labelProb}), any transducer model can be rewritten into a segmental model providing an identical label posterior distribution.  

\subsection{From Segmental Model to Transducer}
Also, we postulate that any segmental model as defined in Sec.~\ref{sbs:segmentalmodel}
can be rewritten into an equivalent transducer model using the RNN-T
topology that also defines exactly the same overall label posterior
distribution. The following transformation equations show
the resulting transducer model written in terms of the segmental
model:\\
\scalebox{0.9}{\parbox{1.11\linewidth}{%
\begin{align*}
  &q(y_{u=t+s-1} \mid y_1^{u-1}, h_1^T) = \numberthis \label{eq:RNNTprob}\\
  &\left\{
    \begin{array}{ll}
      \displaystyle\frac{1-\sum_{\tau=t_{s-1}}^t p(\tau|t_1^{s-1}, a_1^{s-1}, h_1^T)}{1-\sum_{\tau=t_{s-1}}^{t-1} p(\tau|t_1^{s-1}, a_1^{s-1}, h_1^T)} &\text{iff}\;y_u=\blank,\\
      \\
      p(a_s|a_1^{s-1}, t_1^s, h_1^T)\cdot\big(1-q(\blank \mid y_1^{u-1}, h_1^T)\big) \;\; &\text{otherwise} 
    \end{array}
  \right.
\end{align*}}}
with the number of segments $s$, segment labels $a_1^{s}$ and corresponding segment boundaries $t_1^{s}$ defined as follows:\\
\scalebox{0.9}{\parbox{1.11\linewidth}{%
\begin{align*}
  s &=\big|\big\{u'\in \{1,\ldots,u-1\}: y_{u'}\in V\big\}\big| + 1,\\
  t_{s'}&=\min\big\{t\in\{t_{s'-1},\ldots,T\}:y_{t+s'-1}\in V\big\},\\
  a_{s'}&=y_{t_{s'}+s'-1} \quad \forall\;s'=1,\ldots,s-1.
\end{align*}}}
Here $\epsilon$ is again regarded as segment continuation, but the final blank transition at step $U$ is regarded as sentence end label $\#$ in this case. 
Substituting Eq.~(\ref{eq:RNNTprob}) into the transducer model defined
in Eq.~(\ref{eq:transducerSeq}) leads to the segmental model defined
in Eq.~(\ref{eq:segSeq}).
This shows that using Eq.~(\ref{eq:RNNTprob}), also any segmental model can be rewritten
into a transducer model providing an identical label posterior
distribution. Therefore, both segmental model and transducer model are
equivalent and provide the same modeling strength.

\subsection{Special case: strict monotonicity}
\label{sec:mono}
An additional strict monotonicity constraint with a minimum segment length of one frame can be simply adopted by modifying the segment boundary condition to $t_{s-1} < t_s \le T$ and $t_0 \equiv 0$. This can be easily applied into the aforementioned interrelation between segmental model and transducer model by adding the additional constraint:\\
\scalebox{0.9}{\parbox{1.11\linewidth}{%
\begin{align*}
  p(t_s=t_{s-1} \mid a_1^{s-1}, t_1^{s-1}, h_1^T) = 0
\end{align*}}}
which effectively leads to:\\
\scalebox{0.9}{\parbox{1.11\linewidth}{%
\begin{align*}
  q(\blank \mid y_1^{u-1}, h_1^T)  = 1 \; \text{for} \; y_{u-1}\in V.
\end{align*}}}
This corresponds to time-synchronous transducer models \cite{sak2017rna, tripathi2019monoRNNT, zhou2021phonemeTransducer}, where the RNN-T vertical transition is replaced with a diagonal transition, i.e. $u=t$ and $U=T$. 
With this time-synchronous label topology, the given interrelation is still valid. Therefore, such transducer model with strict monotonicity is just one special case and is also equivalently powerful as a segmental model.


\subsection{Search and pruning}
This model equivalence indicates that we can apply the same transducer model as a segmental model in decoding. In general, the final best output sequence can be decided as:\\
\scalebox{0.9}{\parbox{1.11\linewidth}{%
\begin{align*}
x_1^{T'} \rightarrow \tilde{a}_1^{\tilde{S}} = \argmax_{a_1^S, S} p_{\text{LM}}^{\lambda}(a_1^S) \cdot p(a_1^S \mid h_1^T) \numberthis \label{eq:decision}
\end{align*}}}
where the log-linear combination with an external language model (LM) $p_{\text{LM}}$ using scale $\lambda$ is optional. 
Using the Viterbi approximation, \Cref{eq:decision} for the transducer model can be further written as:\\
\scalebox{0.9}{\parbox{1.11\linewidth}{%
\begin{align*}
x_1^{T'} \rightarrow \tilde{a}_1^{\tilde{S}} = \argmax_{a_1^S, S} p_{\text{LM}}^{\lambda}(a_1^S) \cdot \max_{y_1^U:a_1^S} p(y_1^U \mid h_1^T) \numberthis \label{eq:decTransducer}
\end{align*}}}
and for the segmental model as:\\
\scalebox{0.9}{\parbox{1.11\linewidth}{%
\begin{align*}
x_1^{T'} \rightarrow \tilde{a}_1^{\tilde{S}} = \argmax_{a_1^S, S} p_{\text{LM}}^{\lambda}(a_1^S) \cdot \max_{t_1^S} p(a_1^S, t_1^S \mid h_1^T) \numberthis \label{eq:decSegmental}
\end{align*}}}
Ideally, with an identical label posterior distribution and unrestricted decoding conditions, both \Cref{eq:decTransducer} and \Cref{eq:decSegmental} should reveal the same optimal output sequence with the same probability.

\subsubsection{Time-synchronous vs. label-synchronous search}
\label{sec:search}
Although both approaches are equivalent, they result in different search behavior in decoding. More precisely, time-synchronous search is usually used for transducer models \cite{graves2012sequence, tripathi2019monoRNNT, kim2020osc}, where hypotheses $y_1^u$ are expanded per time frame $t$. 
Other variants such as alignment-length synchronous decoding \cite{saon2020ALSD} can also be applied for decoding transducer models, which are not investigated in this work.
On the other hand, segmental models suggest label-synchronous search, where hypotheses $(a_1^s, t_1^s)$ are expanded per segment $s$. This leads to a quadratic cost of search to hypothesize both labels and segment boundaries. Additionally, one can also decompose $(a_1^s, t_1^s)$ to perform search on expansions of $t$ before expanding $a$ or vice versa.

\subsubsection{Pruning}
\label{sec:prune}
Pruning can also be an issue for decoding with such re-interpreted models. A common pruning method applied after each hypotheses expansion is the score-based pruning, where score refers to the negative logarithm of probability.  With score-based pruning, hypotheses are pruned away if their score difference to the current best is more than a predefined threshold $Q_{\text{prune}}$. 
Another simple and common pruning method is to use a fixed beam size $B$. Based on score, only the best $B$ hypotheses at each expansion step are kept for further search.

While these pruning methods are well investigated for time-synchronous search, they may cause more search error for label-synchronous search in this case.
Based on Eqs.~(\ref{eq:durationProb}) and (\ref{eq:labelProb}), sequence hypotheses of the same length might cover completely different number of encoder frames. 
This variation of temporal contribution can lead to an unreliable score comparison between very short and long segment hypotheses before reaching $T$. Thus, search can be expected to become more sensitive to pruning. 
The concrete effect can also vary among different settings such as subsampling and label unit choice.


\section{Experiments}
\subsection{Phoneme-based Transducer on TED-LIUM-v2}
One setup of our experiments is done on the 2nd release of the TED-LIUM corpus (TED-LIUM-v2) \cite{tedlium2}. We use the same phoneme-based transducer model from \cite{zhou2021phonemeTransducer}, which has the strict monotonicity constraint as described in \Cref{sec:mono}. Additionally, the model assumes a first-order dependency which still fits in the equivalence transformation shown in \Cref{sec:t2s}. 
This simplification largely reduces the computation complexity and allows us to investigate the score-based pruning with larger $Q_{\text{prune}}$. We use \Cref{eq:decTransducer} and \Cref{eq:decSegmental} for time-synchronous and label-synchronous search, respectively. 
The same 4-gram word-level LM and scale $\lambda$ as in \cite{zhou2021phonemeTransducer} are used for all recognition. 

In this setup, both time-synchronous and label-synchronous decoders are implemented with the RWTH ASR toolkit \cite{rasr}. The former is performed in the standard way where hypotheses expansion and score-based pruning are applied at each time step. The label-synchronous search is performed on full-segment expansion. Namely, for each partial path $(a_1^{s-1}, t_1^{s-1})$ from step $s-1$, we hypothesize $(a_s, t_s)$ jointly at step $s$ and compute score for this full segment. Then score-based pruning is applied among all new path hypotheses $(a_1^s, t_1^s)$. 
Ended hypotheses, i.e. $t_s=T$, are kept separately for final decision without interfering further search among other paths.

The word error rate (WER) results with various pruning threshold $Q_{\text{prune}}$ are shown in the first three columns of \Cref{tab:werTLv2}. For small $Q_{\text{prune}}$, label-synchronous search suffers a much larger degradation than time-synchronous search, which coincides with the pruning sensitivity mentioned in \Cref{sec:prune}. With an increasing $Q_{\text{prune}}$, the performance of label-synchronous search gets better and better. 
However, due to the smooth distribution of the underlying phoneme-based transducer model, the number of hypotheses explode quickly.
Besides, even with $Q_{\text{prune}}=20$, i.e. a magnitude of $10^9$ in the probability domain, it still does not reach the same WER as time-synchronous search. 
Although the tendency is clear to infer that with further increasing $Q_{\text{prune}}$, both search will eventually give the same performance.
Unfortunately, this leads to a dramatic increase of memory and time for decoding, which can not be performed in this work due to hardware limitation. 
Here the simple label-synchronous search applied has a much worse efficiency for the originally transducer model.
It also indicates the necessity of a more suitable pruning method to match the nature of the model.

To further have some insights on the equivalence perspective, we check the number of utterances where both decodings generate the same transcription (same-trans.) as well as the same transcription with same score (same-score) under a numerical tolerance of $0.0001$.
This is done for $Q_{\text{prune}} \ge 10$ where time-synchronous search output already saturates at the optimum. The results are shown in the last two columns of \Cref{tab:werTLv2}, presented as percentage of the total number of utterances in the dev set. One clear evidence here is the large increment of utterances where both decodings generate the same transcription with exactly the same score. We believe that with further increasing $Q_{\text{prune}}$, this number will eventually approach 100\% as suggested by the model equivalence shown in \Cref{sec:sec2}.


\begin{table}[t!]
\caption{\it WER comparison of time-sync. and label-sync. search using the same phoneme-based transducer model under different pruning threshold; Evaluation on the dev set of TED-LIUM-v2; And percentage of utterances where both search generate the same transcription as well as same transcription with same score}
\centering\label{tab:werTLv2}
\setlength{\tabcolsep}{0.2em}
\begin{tabular}{|c|c|c|c|c|}
\hline
\multirow{2}{*}{$Q_{\text{prune}}$} & \multicolumn{2}{c|}{TED-LIUM-v2 dev WER [\%]} & \multicolumn{2}{c|}{Utterance [\%]} \\ \cline{2-5}
     & {\hspace{1.5mm} time-sync. \hspace{1.5mm}} & label-sync. & same-trans. & same-score \\ \hline 
\hline
4    & 8.6        &  26.4       &     \multirow{3}{*}{-} & \multirow{3}{*}{-} \\
6    & 7.5        &  20.9       &     & \\
8    & 7.2        &  17.5       &     & \\ \hline
10   & 7.1        &  15.7       &     30.2  &   26.0  \\ 
12   & 7.1        &  14.4       &     36.7  &   33.1  \\ 
14   & 7.1        &  13.5       &     42.4  &   37.5  \\ 
20   & 7.1        &  12.4       &     55.8  &   51.9  \\ 
\hline
\end{tabular}
\end{table}

\subsection{BPE-based Transducer on Switchboard}
\label{sec:bpe}
We also perform experiments on the Switchboard corpus \cite{swb} with the byte pair encoding (BPE) \cite{sennrich16bpe} subword-based transducer model from \cite{zeyer2020transducer}. This model uses full context dependency and the time-synchronous label topology as described in \Cref{sec:mono}. In this setup, we apply simple beam search and no additional LM is used. 
Both time-synchronous and label-synchronous decoders are implemented purely in RETURNN \cite{returnn} as a batched fully GPU-based beam search decoder. 
All the code and configuration files are published \footnote{\scriptsize https://github.com/rwth-i6/returnn-experiments/tree/master/2021-segmental-transducer-equivalence}.

For time-synchronous search, hypotheses are expanded at each time frame and a beam size $B$ is applied for pruning.
The label-synchronous search was originally implemented for our hard monotonic latent attention model \cite{zeyer2021latentatt}. 
As described in \Cref{sec:search}, this search is performed on expansions of $t$ first before expanding $a$.
More precisely, at search step $s$ given the $B$ partial path hypotheses $(a_1^{s-1}, t_1^{s-1})_{b=1,...,B}$ from step $s-1$, we firstly hypothesize $t_s$ only and update the score for each $(a_1^{s-1}, t_1^s)_b$ based on \Cref{eq:durationProb}. Then for each $b$, we apply a beam $B_t$ to select the top position hypotheses $(a_1^{s-1}, t_1^s)_b$ based on score, which effectively leads to a total of $B \cdot B_t$ remaining  $(a_1^{s-1}, t_1^s)$. 
Then we expand segment label $a_s$ for each of the remaining $(a_1^{s-1}, t_1^s)$ and update the score for each new hypothesis $(a_1^s, t_1^s)$ based on \Cref{eq:labelProb}.
Finally, the top-scored $B$ hypotheses $(a_1^s, t_1^s)$ are kept for the next search step.
Here the $b$-individual position hypotheses pruning with beam $B_t$ is to avoid search errors based on duration probability only without the label probability yet.

The WER results with different beam settings are shown in \cref{tab:swb:time_label_sync}. In this setup, the performance of time-synchronous and label-synchronous search are much closer, although the former is still slightly better. 
This much smaller difference is probably due to two reasons: Firstly, a high subsampling factor of 6 is applied so that label segments become short in general. Secondly, 
the model produces a much sharper distribution over the BPE units.
Both aspects make the label-synchronous search less sensitive to pruning.
By further increasing the beam sizes $B_t$ and $B$, no improvement is obtained, while the GPU memory is quickly exceeded. 

\begin{table}[t]
\centering
\caption{\it WER comparison of time-sync. and label-sync. search using the same BPE-based transducer model under different beam sizes; Evaluation on the Switchboard Hub5'00 set; Detailed application of search and beam sizes see \Cref{sec:bpe}; Increasing the beam size further did not improve the result in any case}
\setlength{\tabcolsep}{0.5em}
\begin{tabular}{|c|c|c|c|}
\hline
Search & $B_t$ & $B$ & Hub5'00 WER [\%]\\

\hline\hline

\multirow{2}{*}{time-sync.} &  -  & 1 & 14.0\\
& - & 12 & 13.8\\
\hline
\multirow{5}{*}{label-sync.} & 1 & 1 & 14.8\\

 & 4 &  4 & 14.2\\

 & 4 & 12 & 14.0\\

& 100 & 12 & 14.1 \\

& 100 & 32 & 14.1\\
 
 
& 500 & 12 & 14.1 \\
\hline
\end{tabular}
\label{tab:swb:time_label_sync}
\end{table}

\section{Conclusions}
In this work, we proved the equivalence of transducer models and segmental models. We showed transformation equations that allow to rewrite any transducer model into a segmental model and vice versa. This covers both the standard RNN-T topology and the additional strict monotonicity constraint as a special case. Based on this equivalence, both time-synchronous and label-synchronous search can be applied for beam search decoding using either model. 
This is experimentally investigated with the phoneme-based transducer model on TED-LIUM-v2 and the BPE-based transducer model on Switchboard. 
Initial experiments show that standard score-based and beam size-based pruning techniques are insufficient for optimal decoding in a label-synchronous fashion.
Although the equivalence shows that both transducer models and segmental models are equally powerful, a time-synchronous decoding paradigm currently shows more efficient pruning behavior for inherently transducer models. 

Additionally, the transformation equations presented would also allow for time-synchronous decoding of originally segmental models. The analytical investigation of decoding with both transducer and segmental models might inspire future work on improving search and pruning approaches in a more general framework.

\section{Acknowledgements}
This work has received funding from the European Research Council (ERC) under the European Union's Horizon 2020 research and innovation programme (grant agreement No 694537, project ``SEQCLAS'') and partly from a Google Focused Award (``Pushing the Frontiers of ASR: Training Criteria and Semi-Supervised Learning''). The work reflects only the authors' views and none of the funding parties is responsible for any use that may be made of the information it contains.

\bibliographystyle{IEEEtran}
\bibliography{refs}

\end{document}